\documentclass[letterpaper]{article} %
\usepackage{aaai25}  %
\usepackage{times}  %
\usepackage{helvet}  %
\usepackage{courier}  %
\usepackage[hyphens]{url}  %
\usepackage{graphicx} %
\usepackage{natbib}  %
\usepackage{caption} %
\usepackage{algorithm}
\usepackage{algorithmic}

\usepackage{newfloat}
\usepackage{listings}
\usepackage{multirow}
\usepackage{booktabs}
\usepackage{tcolorbox}
\usepackage{color} 
\usepackage{xspace}
\usepackage{amsmath}
\DeclareCaptionStyle{ruled}{labelfont=normalfont,labelsep=colon,strut=off} %
\floatstyle{ruled}
\newfloat{listing}{tb}{lst}{}
\floatname{listing}{Listing}
\newcommand{\methodname}{\textsc{S$^3$c-Math}\xspace}
\newcommand{\datasetname}{\textsc{S$^3$c-MathQA}\xspace}
\newcommand{\fullmethodname}{spontaneous step-level self-correction\xspace}
\newcommand{\tableninept}{\fontsize{9pt}{11pt}\selectfont}

\title{\methodname: Spontaneous Step-Level Self-Correction Makes Large Language Models Better Mathematical Reasoners}
\author {
    Yuchen Yan\textsuperscript{\rm 1,\rm 2}\footnote{Contribution during internship at Meituan Group.},
    Jin Jiang\textsuperscript{\rm 2,\rm 3},
    Yang Liu\textsuperscript{\rm 2},
    Yixin Cao\textsuperscript{\rm 4},
    Xin Xu\textsuperscript{\rm2,\rm 5} \\
    Mengdi Zhang\textsuperscript{\rm 2},
    Xunliang Cai\textsuperscript{\rm 2},
    Jian Shao\textsuperscript{\rm 1}\footnote{Corresponding Author.}
}
\affiliations {
    \textsuperscript{\rm 1}Zhejiang University \\
    \textsuperscript{\rm 2}Meituan Group \\
    \textsuperscript{\rm 3}Peking University \\
    \textsuperscript{\rm 4}Fudan University \\
    \textsuperscript{\rm 5}Hong Kong University of Science and Technology\\
    \{yanyuchen, jshao\}@zju.edu.cn
}

\usepackage{bibentry}

\begin{document}

\maketitle

\begin{abstract}
Self-correction is a novel method that can stimulate the potential reasoning abilities of large language models (LLMs).
It involves detecting and correcting errors during the inference process when LLMs solve reasoning problems. 
However, recent works do not regard self-correction as a spontaneous and intrinsic capability of LLMs. 
Instead, such correction is achieved through post-hoc generation, external knowledge introduction, multi-model collaboration, and similar techniques. 
In this paper, we propose a series of mathematical LLMs called \methodname, which are able to perform \textbf{S}pontaneous \textbf{S}tep-level \textbf{S}elf-\textbf{c}orrection for \textbf{Math}ematical reasoning.
This capability helps LLMs to recognize whether their ongoing inference tends to contain errors and simultaneously correct these errors to produce a more reliable response.
We proposed a method, which employs a step-level sampling approach to construct step-wise self-correction data for achieving such ability. 
Additionally, we implement a training strategy that uses above constructed data to equip LLMs with {\fullmethodname} capacities. 
Our data and methods have been demonstrated to be effective across various foundation LLMs, consistently showing significant progress in evaluations on GSM8K, MATH, and other mathematical benchmarks. 
To the best of our knowledge, we are the first to introduce the \fullmethodname ability of LLMs in mathematical reasoning.

\end{abstract}

\section{Introduction}
Reasoning is one of the essential foundational abilities of large language models (LLMs), showing the capacity of LLMs to tackle complex real-world problems. Nowadays, researchers are increasingly focusing on the performance of LLMs in specific reasoning tasks like mathematics, code, logic, common-sense, etc.~\cite{sun2024survey}. Mathematics is one of the significant branches of reasoning ability of LLMs, and solving a mathematical problem can demonstrate an LLM's ability to decompose, reason, and summarize complex problems. In recent works, the chain-of-thought (CoT) inference~\cite{wei2023chainofthought} has been proven to be a method that can significantly enhance an LLM's ability to solve reasoning problems~\cite{suzgun2022challenging, wang2024chainofthought}. CoT induces the model to explicitly output the reasoning process, gradually deriving a series of intermediate processes or sub-goals, thereby enabling the model to correctly answer reasoning questions. However, during the step-by-step reasoning process, an LLM may still generate errors~\cite{chen2024steplevel,wang2024mathshepherd}. These errors can be propagated to subsequent reasoning processes, leading to incorrect outputs from the LLMs.

\begin{figure}[t]
\centering
\includegraphics[width=0.45\textwidth]{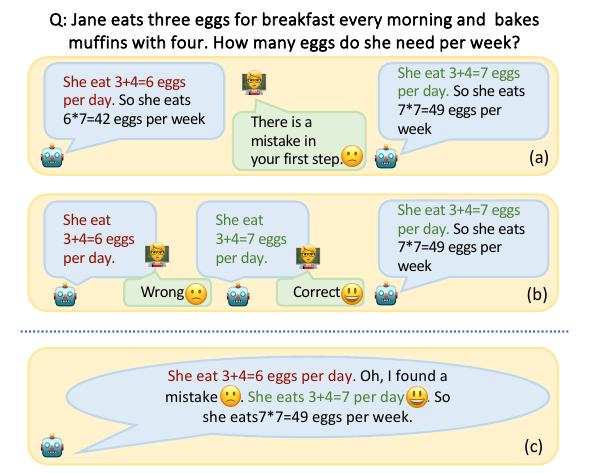} 
\caption{Different ways to implement self-correction. \textbf{(a)} Using more capable LLMs like GPT4 or a trained model to generate the feedback; \textbf{(b)} Apply such process at the step-level; \textbf{(c)} Our work, the LLMs spontaneously identify errors in their output and correct them immediately. \textcolor[RGB]{136,33,16}{Red} text denotes the wrong step, and \textcolor[RGB]{94,129,63}{green} text denotes the content after correction. }
\label{fig1}
\end{figure}

To alleviate potential errors that may occur during model reasoning, self-correction methods have been proposed. These methods can check whether the model makes mistakes during the reasoning process, pinpoint the errors, and enable the model to generate a better answer based on the provided feedback~\cite{kamoi2024when, pan2024automatically}. Existing research has already leveraged such error correction processes to make the model's response more reliable\cite{puerto2024finetuning, paul2024refiner, xu2024pds}. The crux of self-correction lies in \textbf{how to generate feedback} for the model's output, and \textbf{how to generate a better response} based on this feedback~\cite{tyen2024llms}. 

In terms of generating feedback, for some specific tasks, external tools or knowledge can be used as feedback. For instance, in the task of code generation, the code generated by the model can be handed over to the compiler. If the code cannot compile, the model can be informed that an error exists, and the error stack can be provided to the LLMs~\cite{zheng2024opencodeinterpreter,li2024dotamath}. For tasks that are difficult to judge with external tools, some research provides the model's original answer to more capable models such as GPT4 or a trained critic model specifically for error detection, thereby generating corresponding feedback~\cite{zhang2024small}. Such self-correction is usually applied at the instance-level, i.e., after the model's complete output, feedback generation and error correction are completed.
Given that most reasoning tasks are solved in a CoT manner, these self-correction methods have begun to be applied at the step-level, providing the model with more granular signals. We illustrate these self-correction methods in panels (a) and (b) of Figure \ref{fig1}.

However, we believe that the aforementioned methods still seem unnatural in LLMs' inference. 
Utilizing multiple models to address a single problem does not provide a fair assessment of the LLM's reasoning ability, and such an approach cannot be considered an end-to-end capability of the model. Therefore, some researchers have begun to propose implementing all steps of self-correction within the same model, using multi-task learning to enable a single model to master both problem-solving and error correction tasks~\cite{madaan2023selfrefine,liu2024large}. Yet, we do not think this is enough. %
Multi-stage processing requires LLMs to generate several times, thus increasing inference time and computational costs. 
Moreover, this behavior is not spontaneous from the model itself, but a preset fixed process in multi-stage processing, which cannot demonstrate the model's capacity for error recognition.
Based on these considerations, we propose \textbf{\fullmethodname} capability for LLMs, which allows the LLM to spontaneously identify errors in its on-going output and correct them immediately, ultimately yielding more reliable responses. This process is illustrated in panel (c) of Figure~\ref{fig1}.

In this paper, we propose an ingenious method for constructing self-correction data to achieve \fullmethodname. This method utilizes existing step-by-step reasoning instruction data, generating potentially erroneous steps through sampled generation, and validating whether there are indeed errors in the steps through the $pass@k$ validation. 
We insert the erroneous steps and the markers used to trigger self-correction into the correct steps in the existing data and construct \datasetname, which includes 532K self-correction data, based on MetaMathQA~\cite{yu2024metamatha}. 
In order to make self-correction more accurate and generalize to more mathematical problems, we annotate each correction case in a fine-grained manner, adding the reflection and improvements. 
During the training stage, we mix \datasetname and the original 395K MetaMathQA to create 927K SFT data, and use loss-masks to ignore the loss of the erroneous steps. 
This method ensures that the LLMs are introduced with a new self-correction capability while maintaining their original effectiveness. 
Our experiments show stable and consistent improvements across multiple mathematical datasets such as GSM8K~\cite{cobbe2021training} and MATH~\cite{hendrycks2021measuring} on both generalist LLMs like Meta-Llama-3-8B ~\cite{metaai2024introducing} and meth-specialized LLMs like DeepSeek-Math~\cite{shao2024deepseekmath}.
Our main contributions are as follows:

\begin{itemize}
    \item 
    We introduce \fullmethodname capability of LLMs, which can automatically correct potential errors and generate a more reliable response in the reasoning process. To the best of our knowledge, we are the first to propose this capability in mathematical reasoning of LLMs.
    \item
    We propose a method for constructing \fullmethodname data, which could be used to quickly and effectively build self-correction data from existing step-by-step SFT data. With the proposed method, we have constructed \datasetname, a 532K SFT dataset to produce the ability of \fullmethodname.
    \item
    We employ a fine-tuning strategy incorporating loss-mask on the \datasetname, resulting in the training of \methodname, a series of models with \fullmethodname capabilities in mathematical reasoning. This strategy enables the LLMs to identify potential errors during the output process and autonomously correct them. The trained \methodname, with this capability, achieves performance improvements on multiple mathematical benchmarks.
\end{itemize}

\begin{figure*}[!ht]
\centering
\includegraphics[width=0.9\textwidth]{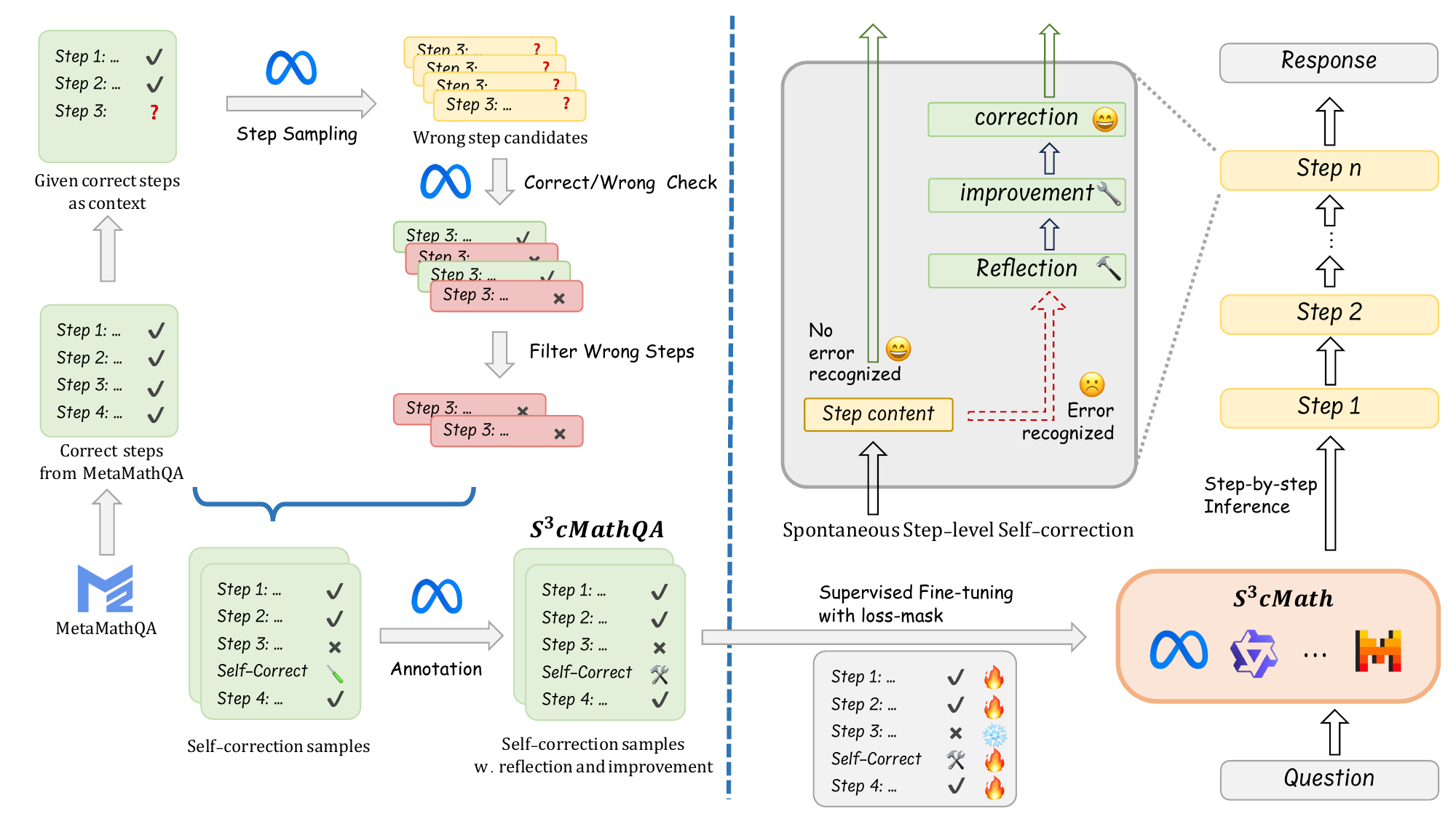} %
\caption{An overview of our works. The left part of this figure illustrates the process of constructing \datasetname, and the right part depicts the training of \methodname and the procedure during its reasoning.}
\label{fig2}
\end{figure*}

\section{Related Works}
\subsection{Mathematical Reasoning of LLMs}
Mathematical reasoning, one of the foundational abilities of LLMs, demonstrates the model's capability to solve complex real-world mathematical problems, which can be enhanced during various stages of LLMs' training. 
In the pre-training stage, continuing pre-training on large amounts of mathematical corpora can improve the internal mathematical reasoning ability of LLMs from a knowledge perspective~\cite{ying2024internlmmath,shao2024deepseekmath, zhou2024jiuzhang3}. 
During supervised instruction tuning, mathematical abilities could be further enhanced by training on problem-answer pairs, leading to the same answer formatting and better reasoners \citep{yu2024metamatha, li2024xwinmath}.
For the preference alignment stage, preference optimization~\cite{chen2024steplevel} and reinforcement learning~\cite{lightman2023let,wang2024mathshepherd} are used to improve LLMs at a fine-grained level. 

\subsection{Self-Correction of LLMs}
Self-correction is an advanced technique to enhance an LLM's output by refining its initial response during the inference process, which can be implemented through various methodologies~\citep{kamoi2024when, pan2024automatically}.
Strategies of self-correction center around the concept of feedback generation. Essentially, the LLM must evaluate whether errors exist in its initial output, identify these errors, and pinpoint the steps needed for correction. The timing of self-correction can be categorized into two main types: post-hoc and real-time~\citep{paul2024refiner, jiang2023active}. 
In the post-hoc approach, feedback is generated after the initial inference, based on which the LLM is prompted to re-generate the response.
In contrast, the real-time approaches integrate feedback into the generation context during the inference process, enabling the LLM to produce a completion with the necessary corrections on the fly. 
Additionally, feedback of the LLMs' contents could be generated in two types: external knowledge and tools to such as code compilers~\citep{zheng2024opencodeinterpreter,li2024dotamath} or LLMs themselves.
Feedback generation by LLMs can be bifurcated into cross-model and same-model approaches. In the cross-model approach, a more advanced LLM or a dedicatedly trained critic model is employed to produce the feedback~\cite{li2024prd,cohen2023lm,du2023improving}. On the other hand, the same-model approach utilizes the same LLM for both the initial generation and the feedback, albeit with different prompts to guide the correction process~\citep{thorne2021evidencebased,tyen2024llms,paul2024refiner}.

\subsection{Synthesised Mathematical Data}
Synthesized data generated from existing LLMs has been proven effective for training another LLM~\cite{yu2024metamatha, liu2024mmiqc, li2024xwinmath, zhu2024kpmath, xu2024extension, zhou2024jiuzhang3}, which can be used in both pre-training and supervised fine-tuning.
For pre-training, data can be synthesized to expand the scope of mathematical knowledge that LLMs can learn from. For instance, continuing pre-training on the synthesized Jiuzhang dataset~\cite{zhou2024jiuzhang3} significantly improves LLMs' performance on multiple mathematical benchmarks. %
For supervised fine-tuning, there are also various methods to synthesize supervised instruction pairs. 
For example, MetaMathQA~\cite{yu2024metamatha} augments questions from the original training set and generates answers for those bootstrapped questions. More recently, KPMath~\cite{zhu2024kpmath} extracts key points from existing data to analyze and synthesize more complex mathematical word problems.

\section{Approach}
The main works of this paper can be divided into two parts. The first part is to construct self-correction SFT data \datasetname based on existing step-by-step mathematical instruction pairs. The second is to use \datasetname to carry out instruction fine-tuning, ultimately enabling the model to acquire \fullmethodname capabilities.

\subsection{Data Construction}
We divide our proposed self-correction data construction process into two steps. The first step involves generating erroneous steps from existing correct CoT steps, which will be used to create step-level self-correction instances. The second step concerns reflections and improvements for the above step-level self-correction instances, which serve as detailed guidance for LLMs to better acquire the ability to do \fullmethodname. In this paper, we use the MetaMathQA~\citep{yu2024metamatha} dataset, which includes 395K mathematical reasoning step-by-step samples, as the seed dataset.

\subsubsection{Wrong Step Sampling}

Before we start, we carry out instruction tuning on the Meta-Llama-3-8B~\citep{metaai2024introducing} using the MetaMathQA data, resulting in our model $M$ used for sampling erroneous steps. Since we need to sample on the training set, in order to mitigate the reduction of diversity caused by model over-fitting, we adopt the idea of cross-validation and conduct a 5-fold cross-training. That is, we evenly divide the MetaMathQA data into five parts, use four parts of the data to train the model, and the remaining one part of the data for sampling erroneous steps. This process is repeated five times on different parts of the data. Hereafter, we denote the data used for model training as $s$, and the remaining data used for sampling erroneous steps as $\bar{s}$. The model that undergoes instruction tuning on $s$ is denoted as $M_s$.

First, we use the trained model $M_s$ to perform step-wise sampling on the split-out data $\bar{s}$, obtaining a set of steps that may contain errors. The specific method is that for each step $x_i$ in an SFT sample response $x$, we concatenate all the steps from the first to the $i^{th}$ step into a context and give it to $M_s$ for continuation, until the model outputs the next step, obtaining $candidates(\hat{x}_{i+1})$. We use a high temperature and sample multiple steps at the same time to ensure the diversity of steps. The candidates of the $(i+1)^{th}$ steps of case $x$ could be represented as:
$$ candidates(\hat{x}_{i+1}) = M_{s}^{k}(x_1\oplus\ldots\oplus x_i) , x \in \bar{s} $$
where $\oplus$ indicates concatenation of existing correct steps, and $M_{s}^{k}(x_1\oplus\ldots\oplus x_i)$ represents the set of responses generated by $M_s$ for $x_1\oplus\ldots\oplus x_i$ k times.

Next, we will use the trained model to evaluate the correctness of the generated candidates $(\hat{x}_{i+1})$. Specifically, for each step $\hat{x}_{i+1}$ in the candidate set, we concatenate the correct preceding steps $x_1,\ldots,x_i$ as context before the step $\hat{x}_{i+1}$, and ask the model $M_s$ to continue generating until the model actively ends the generation (i.e., encounters the [end-of-sequence] token). We will determine whether the step is correct based on whether the answer generated by model $M_s$ matches the standard answer. In order to make more accurate judgments and reduce the cases where correct steps are mistakenly considered as wrong, we use $pass@k$ to evaluate a step. Specifically, we sample multiple model outputs with a high temperature of 1.0 for 16 times, and only when none of the contents generated by the model match the correct answer do we consider the step to be wrong. Whether $\hat{x}_{i+1}$ is correct can be represented as:
$$
\fontsize{8.5pt}{11pt}\selectfont
\hat{x}_{i+1}
\begin{cases} 
\text{Correct}&\text{if } \exists o \in M_s^k(x_1\oplus...\oplus x_i \oplus \hat{x}_{i+1}), A(o) = A(gt) \\
\text{Wrong}&\text{if } \forall o \in M_s^k(x_1\oplus...\oplus x_i \oplus \hat{x}_{i+1}), A(o) \neq A(gt) 
\end{cases}
$$
where $A(o)$ represents the answer extraction for the output $o$, while $A(gt)$ stands for the answer extraction for the golden truth.

\begin{figure}[h]
\centering
\includegraphics[width=0.45\textwidth]{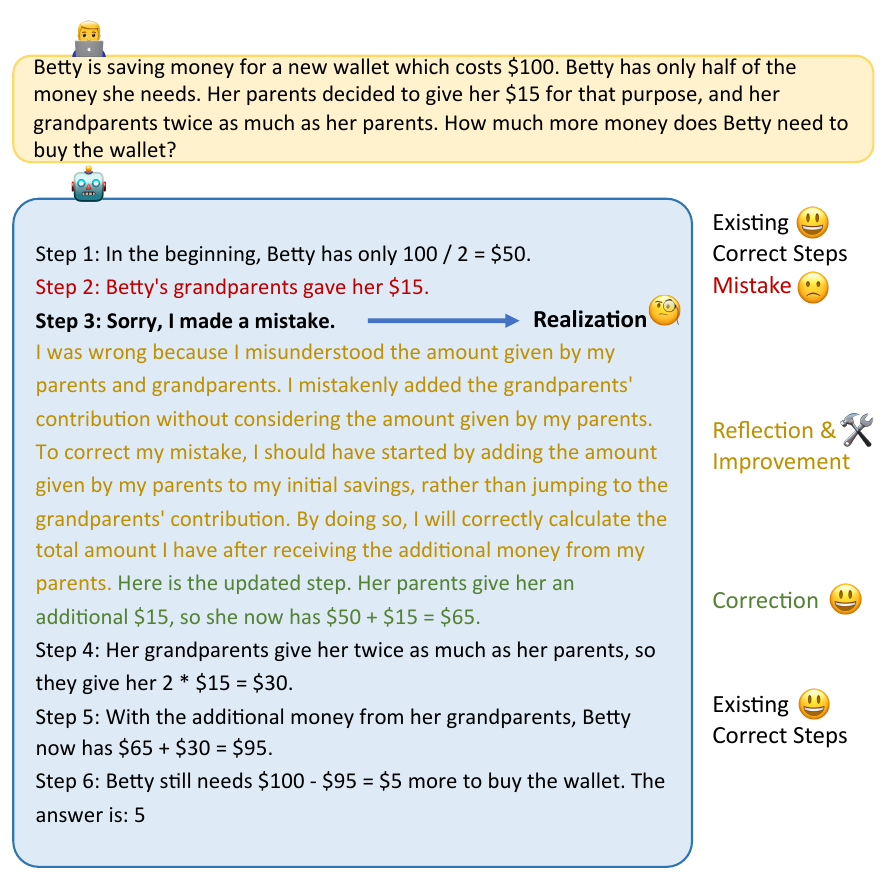} %
\caption{An example from \datasetname. Steps 1, 4, 5, and 6 are copied from the correct steps already exist in MetaMathQA. Step 2 is a wrong step sampled from the method we proposed. Step 3 is the indicator of LLM's error realization, reflection, improvement, and the new corrected step. During training, we do not compute the loss for the \textcolor[RGB]{136,33,16}{red} texts.}
\label{fig3}
\end{figure}

\subsubsection{Reflection and Improvement Generation}

Based on the correct steps in the existing data and the erroneous steps we sampled in the previous section, we inserted the erroneous steps and the flags used to trigger self-correction into the correct steps, thus forming the direct self-correction data. However, another significant challenge of self-correction is how the LLMs can produce more accurate content when they know that errors already exist. To enable the LLMs to better learn to correct the identified mistakes, we annotated the acquired step-level self-correction samples in two ways, reflection and improvement, using Meta-Llama-3-70B-Instruct~\citep{metaai2024introducing}. Reflection involves the model analyzing where errors occurred in the existing answers, while improvement entails generating ways to improve based on the existing output and reflection. When generating these two annotations, we provide the model with a prompt, as well as a question, previous steps, wrong step, and correct step, and ask the model to make annotations based on this information. We present a data example and its components in Figure \ref{fig3}.

\begin{table*}[t]
\centering
\setlength{\tabcolsep}{1.2mm}
\begin{tabular}{rlrrrrrrrr}
\toprule
\multicolumn{1}{l}{\multirow{2}[4]{*}{\textbf{Base Model}}} & \multirow{2}[4]{*}{\textbf{SFT Data}} & \multicolumn{2}{c}{GSM8K} & \multicolumn{2}{c}{MATH} & \multicolumn{2}{c}{SVAMP} & \multicolumn{2}{c}{Mathematics} \\
\cmidrule{3-10}      &       & \multicolumn{1}{c}{~$P@1$} & \multicolumn{1}{c}{$M@32$} & \multicolumn{1}{c}{~$P@1$} & \multicolumn{1}{c}{$M@32$} & \multicolumn{1}{c}{~$P@1$} & \multicolumn{1}{c}{$M@32$} & \multicolumn{1}{c}{~$P@1$} & \multicolumn{1}{c}{$M@32$} \\
\midrule
\multicolumn{10}{c}{Generalist Models} \\
\midrule
\multicolumn{1}{l}{Meta-Llama-3-8B} & MetaMath & 81.12  & 86.20  & 30.58  & 39.92  & \underline{81.40}  & \textbf{86.60} & \underline{18.27}  & \textbf{28.21} \\
\multicolumn{1}{l}{\cite{metaai2024introducing}} & \quad + \datasetname - w/o. R\&I & \underline{81.65}  & \textbf{88.10} & \underline{32.32}  & \underline{41.00}  & 80.50  & 84.60  & 17.56  & 25.48  \\
      & \quad + \datasetname & \textbf{82.94} & \underline{87.34}  & \textbf{33.14} & \textbf{41.60} & \textbf{81.80} & \underline{86.40}  & \textbf{19.08} & \underline{27.53}  \\
\midrule
\multicolumn{1}{l}{Mistral-7B-v0.3} & MetaMath & 73.46  & \underline{82.34}  & 24.04  & 31.34  & \underline{77.20}  & 83.30  & \textbf{17.35} & \underline{27.98}  \\
\multicolumn{1}{l}{\cite{mistralaiteam2023mistral}} & \quad + \datasetname - w/o. R\&I & \underline{75.36}  & 81.80  & \textbf{27.30} & \underline{32.28}  & 75.60  & \underline{83.40}  & 13.48  & 25.89  \\
      & \quad + \datasetname & \textbf{75.51} & \textbf{82.64} & \underline{25.48}  & \textbf{33.74} & \textbf{78.40} & \textbf{84.90} & \underline{16.07}  & \textbf{28.10} \\
\midrule
\multicolumn{1}{l}{Meta-Llama-3-70B} & MetaMath & 88.55  & 91.74  & \underline{45.70}  & \underline{55.40}  & 87.40  & \textbf{91.00} & \underline{29.08}  & \underline{40.30}  \\
\multicolumn{1}{l}{\cite{metaai2024introducing}} & \quad + \datasetname - w/o. R\&I & \underline{89.16}  & \underline{92.72}  & 44.16  & 53.46  & \underline{87.60}  & \underline{90.70}  & 28.48  & 39.02  \\
      & \quad + \datasetname & \textbf{91.66} & \textbf{93.33} & \textbf{46.22} & \textbf{55.80} & \textbf{87.70} & \textbf{91.00} & \textbf{34.17} & \textbf{42.65} \\
\midrule
\multicolumn{10}{c}{Math-specialized Models} \\
\midrule
\multicolumn{1}{l}{Deepseek-math-base} & MetaMath & 79.30  & 85.90  & 38.22  & 46.86  & 80.60  & \underline{86.40}  & \underline{25.98}  & 42.95  \\
\multicolumn{1}{l}{\cite{shao2024deepseekmath}} & \quad + \datasetname - w/o. R\&I & \underline{82.18}  & \underline{87.72}  & \underline{40.08}  & \underline{50.92}  & \underline{81.80}  & \underline{86.40}  & 23.75  & \underline{43.93}  \\
      & \quad + \datasetname & \textbf{82.49} & \textbf{88.17} & \textbf{41.40} & \textbf{52.14} & \textbf{82.20} & \textbf{87.60} & \textbf{28.27} & \textbf{45.68} \\
\midrule
\multicolumn{1}{l}{Qwen2-Math-7B} & MetaMath & 84.08  & 89.08  & 51.32  & 60.64  & 85.60  & \textbf{90.40} & \textbf{42.44} & \underline{52.02}  \\
\multicolumn{1}{l}{\cite{qwenteam2024introducing}} & \quad + \datasetname - w/o. R\&I & \underline{84.38}  & \underline{89.39}  & \textbf{51.94} & \textbf{62.80} & \underline{86.70}  & \underline{89.50}  & \underline{39.73}  & 51.61  \\
      & \quad + \datasetname & \textbf{84.76} & \textbf{89.61} & \underline{51.76}  & \underline{62.40}  & \textbf{87.40} & 89.40  & 39.35  & \textbf{52.20} \\
\midrule
\multicolumn{1}{l}{Qwen2-Math-72B} & MetaMath & 88.86  & \underline{91.81}  & \underline{54.66}  & \underline{64.68}  & \underline{87.80}  & \underline{91.10}  & \underline{45.36}  & \textbf{59.17} \\
\multicolumn{1}{l}{\cite{qwenteam2024introducing}} & \quad + \datasetname - w/o. R\&I & \underline{88.40}  & 91.74  & 53.76  & 64.06  & 87.50  & 91.00  & 44.55  & 58.54  \\
      & \quad + \datasetname & \textbf{88.93} & \textbf{93.10} & \textbf{55.12} & \textbf{66.22} & \textbf{89.30} & \textbf{91.50} & \textbf{46.28} & \underline{58.93}  \\
\bottomrule
\end{tabular}%
\caption{Our main experimental results (\%) on four mathematical reasoning tasks (GSM8K, MATH, SVAMP and Mathematics) under $Pass@1$ and $Majority@32$ settings. The term "w/o. R\&I" represents that the self-correction part of the SFT samples do not include reflection and improvements. We abbreviate $Pass@1$ as $P@1$ and $Majority@32$ as $M@32$. The highest accuracy is indicated in \textbf{bold}, while the second highest is \underline{underlined}.
}
\label{main_results}
\end{table*}

\subsection{Tuning and Evaluation}
After constructing the aforementioned data, we implement SFT on it. We test our trained model on multiple mathematical evaluation benchmarks to validate the effectiveness of our data and methods.
\subsubsection{Instruction Tuning}
We conduct SFT on the constructed dataset. Since our SFT data is self-correction data, it contains at least one erroneous step. Learning from these erroneous steps might harm the reasoning performance of the model itself. To avoid being affected by this situation, we use a loss mask to control the learning objectives of the model. Specifically, we applied a mask operation to the erroneous step, preventing the model from learning this error, thereby maintaining the performance of the existing SFT data. As shown in the red section of Figure \ref{fig3}, we will not calculate the loss for it.

\subsubsection{Evaluation Settings}

We test the performance of our models on the evaluation sets of GSM8K~\citep{cobbe2021training}, MATH~\citep{hendrycks2021measuring}, SVAMP~\citep{patel2021are}, and Mathematics~\citep{saxton2019analysing}. Each test sample in these datasets contains a question and a golden answer. We extract the final answer from the model's output and matched it with the golden answer to ultimately determine whether the sample was answered correctly. The evaluation uses the \textit{vllm} framework ~\cite{kwon2023efficient} for inference, and we evaluate the model's single-inference accuracy, i.e., the $pass@1$ metric, using greedy decoding. Self-consistency~\citep{wang2023selfconsistency} is considered a stable way to evaluate the model's sampling method. To more accurately assess the improvements brought about by our strategy, we also test the accuracy under the $maj@32$ setting, setting the temperature to 0.7 during sampling decoding.

\section{Experiments}
\subsection{Baseline}
Since our data is built on MetaMathQA\cite{yu2024metamatha} dataset, we use the model fine-tuned on MetaMathQA as the baseline for our experiments. We compare the performance difference between our trained model and the baseline model on several mathematical benchmarks. All our SFT experiments use the same configuration. We use Megatron-LM as the training framework and set the initial learning rate to 1e-5, but except 1e-6 for Mistral-7B-v0.3\cite{mistralaiteam2023mistral}, with a 0.03 warm-up ratio, and decay to 0 using a cosine schedule. The model's max length is 8192, the global batch size is 128, and the number of training epochs is 3. We pack all SFT data to accelerate the training process. We conducted our SFT experiments on 32 Nvidia A100 GPUs.

\subsection{Main Results}

Based on MetaMathQA 395K, We applied our proposed method to synthesize \datasetname, 532K self-correction data. For all instances in \datasetname, we generated reflection and improvement using our proposed methods. The main motivation of our paper is to seamlessly introduce the end-to-end intrinsic self-correction capability into the model. Therefore, we merged the original MetaMathQA dataset with our synthesized data \datasetname, yielding a total of 927K data for our SFT data corpus.

To demonstrate the generalization effect of our data, we conducted SFT experiments on different foundation LLMs using the MetaMathQA and \datasetname. We adopted both generalist LLMs and math-specialized LLMs. For the generalist LLMs, we utilized three different size base models: Meta-Llama-3-8B\cite{metaai2024introducing}, Mistral-7B-v0.3\cite{mistralaiteam2023mistral}, and Meta-Llama-3-70B\cite{metaai2024introducing}. For the math-specialized models, we used the DeepSeek-Math\cite{shao2024deepseekmath}, Qwen2-Math-7B \cite{qwenteam2024introducing} and Qwen2-Math-72B\cite{qwenteam2024introducing}.

We evaluated the trained model on GSMK, MATH, SVAMP, and Mathematics, with the results shown in Table \ref{main_results}. Our experimental results demonstrate that our model can make stable improvements across multiple mathematical evaluation benchmarks. On Meta-Llama-3-8B, we improved the accuracy of GSM8K from 81.1\% to 82.9\%, and the accuracy of MATH from 30.6\% to 33.1\%. On Deepseek-Math-Base, we increased the accuracy of GSM8K from 79.3\% to 82.5\%, and the accuracy of MATH from 38.2\% to 41.4\%. Our method achieved consistent improvements on both the generalist LLMs and the math-specialized LLMs, thereby validating the effectiveness of the method we proposed.

In our experimental results, we compared three methods: direct correction, correction after reflection, and correction after reflection and generation of improvement. The results showed that on multiple datasets, training the model with fine-grained analysis information can enhance the LLM's performance in mathematical abilities.

\section{Analysis}
\subsection{Ablation Study}

\subsubsection{Distribution Change vs. Self-Correction Introduction}

Since we introduced new data into the original MetaMathQA dataset, even though our data's queries all come from MetaMathQA, there is still a possibility of changing the query distribution of the SFT data. Our error path sampling method may generate more self-correction data on more challenging problems, thus more difficult queries may appear more frequently in our SFT data. To ablate the impact of these situations on our results and to prove that our improvement comes from self-correction itself rather than the distribution changes caused by sampling, we conducted the following ablation experiment. The specific method of this experiment is that, based on the queries of the 532K self-correction data we generated, we over-sampled the original data from MetaMathQA, resulting in a total of 927K data. The distribution of queries in the over-sampled data is the same as that in our proposed 927K data.

\begin{table}[h]
  \centering
  \setlength{\tabcolsep}{1mm}
  \tableninept
\begin{tabular}{lrrrr}
\toprule
\multicolumn{1}{c}{\multirow{2}[4]{*}{\textbf{SFT Data}}} & \multicolumn{2}{c}{\textbf{GSM8K}} & \multicolumn{2}{c}{\textbf{MATH}} \\
\cmidrule{2-5}      & \multicolumn{1}{c}{$P@1$} & \multicolumn{1}{c}{$M@32$} & \multicolumn{1}{c}{$P@1$} & \multicolumn{1}{c}{$M@32$} \\
\midrule
MetaMathQA & 81.12  & 86.20  & 30.58  & 39.92  \\
\quad+ \datasetname - w/o. R\&I & \textbf{81.65}  & \textbf{88.10}  & \textbf{32.32}  & \textbf{41.00}  \\
MetaMathQA dist-aligned & 78.54  & 85.22  & 30.92  & 38.62  \\
\bottomrule
\end{tabular}%

  \caption{Data Distribution Change vs. Self-correction Introduction. "MetaMathQA dist-aligned" represents data that has the same distribution as the query in "MetaMathQA + \datasetname", but uses the response from MetaMathQA.}
  \label{tab:table2}%
\end{table}%

As shown in Table \ref{tab:table2}, altering the distribution of the query does not enhance the model's performance, further validating the effectiveness of our method.

\begin{table*}[htbp]
\setlength\tabcolsep{4pt}
\scriptsize

\centering
\begin{tabular} {p{2cm}|p{15cm}}
        \toprule
         \textbf{Question} & \textit{Cecilia just bought a new puppy. According to her veterinarian, she has to feed the puppy 1 cup of dog food every day for the first 180 days. Then she has to feed the puppy 2 cups of dog food every day for the rest of its life. If one bag of dog food contains 110 cups, how many bags of dog food will Cecilia use in the first year?}
         \\
         \midrule
         \multirow{3}{*}{w/o. self-correction} & Step 1: In the first 180 days, Cecilia will feed the puppy 1 cup of dog food every day, so she will use 180 cups of dog food. \\
& Step 2: In the rest of the puppy's life, she will feed the puppy 2 cups of dog food every day, so she will use 2 cups x 365 days = 730 cups of dog food.\\
&Step 3: In total, Cecilia will use 180 cups + 730 cups = 910 cups of dog food in the first year.\\
&Step 4: Since one bag of dog food contains 110 cups, Cecilia will need 910 cups / 110 cups per bag = 8.27 bags of dog food.\\
&Step 5: Since she cannot buy a fraction of a bag, she will need to round up to the nearest whole number.\\
&Step 6: Therefore, Cecilia will use 9 bags of dog food in the first year. The answer is: 9\\
         \midrule
         \multirow{6}{*}{w. self-correction} & Step 1: In the first 180 days, Cecilia will need to feed the puppy 1 cup of dog food every day, so she will need 1 cup x 180 days = 180 cups of dog food. \\
& Step 2: In the rest of the puppy's life, she will need to feed the puppy 2 cups of dog food every day, so she will need 2 cups x 365 days = 730 cups of dog food. \\
& \textcolor[RGB]{136,33,16}{Step 3: Sorry, I made a mistake. I incorrectly assumed that the rest of puppy's life would be 365 days, which is the number of days in a year, and used that as the number of days to calculate the total cups of dog food needed for the rest of the puppy's life. However, the problem states that the rest of its life is not just 365 days.} \\
& \textcolor[RGB]{136,33,16}{To correct this, I should have used a variable to represent the number of days the puppy will live after the first 180 days, and then multiplied that variable by 2 cups to get the total cups of dog food needed for the rest of the puppy's life. This will allow me to consider the possibility that the puppy may live for a longer or shorter period of time. Here is the updated step. In the rest of the puppy's life, she will need to feed the puppy 2 cups of dog food every day, so she will need 2 cups x (365 days - 180 days) = 370 cups of dog food.} \\
& Step 4: In total, Cecilia will need 180 cups + 370 cups = 550 cups of dog food in the first year. \\
& Step 5: Since one bag of dog food contains 110 cups, Cecilia will need 550 cups / 110 cups per bag = 5 bags of dog food in the first year. The answer is: 5. \\
         \midrule
    \end{tabular}
    \caption{A case study from the GSM8K test dataset. \textcolor[RGB]{136,33,16}{Red} text denotes the process of \fullmethodname.}
    \label{tab:case}
\end{table*}

\subsubsection{Step-Level Correction vs. Instance-Level Correction}

The method we propose is a step-level self-correction, the advantage of which is that the model can directly correct the errors just occurred in the output process in a timely manner, thereby avoiding unnecessary error propagation. We can also generalize self-correction case-wise, that is, allowing the model to output all the content at once, and then let the model probe the generated content for errors. If errors are found, the model will automatically do self-correction, thereby producing a more accurate answer. To prove that the step-level self-correction in our work is more effective, we carried out this ablation experiment. Specifically, we saved the complete model output used to verify the error step at the error step sampling stage. According to our method, the answer paths sampled here cannot ultimately derive the correct answer. We take the entire error path as the context to generate instance-level correction data. Consistent with our proposed method, we also do not calculate loss on the erroneous reasoning path.

\begin{table}[h]
  \centering
    \tableninept
    \setlength{\tabcolsep}{1mm}
    \begin{tabular}{lrrrr}
    \toprule
    \multicolumn{1}{c}{\multirow{2}[4]{*}{\textbf{SFT Data}}} & \multicolumn{2}{c}{\textbf{GSM8K}} & \multicolumn{2}{c}{\textbf{MATH}} \\
    \cmidrule{2-5}      & \multicolumn{1}{c}{$P@1$} & \multicolumn{1}{c}{$M@32$} & \multicolumn{1}{c}{$P@1$} & \multicolumn{1}{c}{$M@32$} \\
    \midrule
    MetaMathQA & 81.12  & 86.20  & 30.58  & 39.92  \\
    \quad+ instance-level correction & 79.98  & 87.26  & 32.22  & 40.60  \\
    \quad + \datasetname - w/o. R\&I & \textbf{81.65}  & \textbf{88.10}  & \textbf{32.32}  & \textbf{41.00}  \\
    \bottomrule
    \end{tabular}%

  \caption{Step-Level correction vs. Instance-level correction. "instance-level correction" utilized the same sample size for training as \datasetname.}
  \label{table3}%
\end{table}%

As shown in Table \ref{table3}, experimental results indicate that while instance-level self-correction can yield some improvement, the extent of this enhancement is not as significant as that achieved with the step-level approach we employed.

\subsubsection{Generate From Existing Paths vs. Generate New Paths}

One of the key features of our method is to directly construct self-correction data from the existing step-by-step mathematical data. Therefore, in our correction data, apart from the erroneous step and the corresponding correction behavior, the remaining correct steps are completely consistent with the original SFT data. This allows us to seamlessly introduce the intrinsic self-correction ability into the LLMs. Currently, in exploring the mathematical capabilities of the LLM, scholars use Monte Carlo Tree Search (MCTS) for step sampling~\cite{wang2024mathshepherd, zhang2024restmcts}, which allows for more diverse steps and decouples from the existing SFT data. To demonstrate that our direct sampling of erroneous steps from existing SFT data and construction of self-correction data is more effective, we designed this ablation experiment. We implemented an MCTS to generate correct and incorrect steps for all queries in MetaMathQA, and used these steps to construct self-correction data. For a fair comparison, we also included the original MetaMathQA data and constructed MCTS self-correction data of the same scale as the data volume as \datasetname.

\begin{table}[h]
  \centering
\setlength{\tabcolsep}{1mm}
\tableninept
\begin{tabular}{lrrrr}
\toprule
\multicolumn{1}{c}{\multirow{2}[4]{*}{\textbf{SFT Data}}} & \multicolumn{2}{c}{\textbf{GSM8K}} & \multicolumn{2}{c}{\textbf{MATH}} \\
\cmidrule{2-5}      & \multicolumn{1}{c}{$P@1$} & \multicolumn{1}{c}{$M@32$} & \multicolumn{1}{c}{$P@1$} & \multicolumn{1}{c}{$M@32$} \\
\midrule
MetaMathQA & 81.12  & 86.20  & 30.58  & 39.92  \\
\quad+ MCTS corrections & 79.98  & 86.96  & 30.96  & 40.12  \\
\quad+ \datasetname - w/o. R\&I & \textbf{81.65}  & \textbf{88.10}  & \textbf{32.32}  & \textbf{41.00}  \\
\bottomrule
\end{tabular}%
  \caption{Generate from existing paths vs. Generate new paths. "MCTS corrections" utilized the same sample size for training as \datasetname.}
  \label{table4}%
\end{table}%

As shown in Table \ref{table4}, the experiments have demonstrated that the step-level self-correction data generated by MCTS cannot achieve the effectiveness of our method. The data produced by our method is more closely aligned with the original SFT data, enabling the introduction of \fullmethodname capabilities within the model while preserving the original SFT data's effectiveness.

\subsection{Case Study}

To more clearly demonstrate the \fullmethodname ability generated by our method in LLMs, we provide an evaluation output o25
f a model trained on the DeepSeek-Math base. We present the results of greedy decoding produced by training on MetaMathQA and \datasetname respectively in Table~\ref{tab:case}. We can see that after training on \datasetname, the model spontaneously realizes its mistake in the second step where it incorrectly assumed `the rest of its life` to represent 365 days, and proceeds to recalculate, arriving at the final result. This case can also show that our method does not affect the original data to the greatest extent, as the models trained on these two sets of data produce almost identical greedy outputs on the same case in the first two steps.

\section{Conclusion}

In this paper, we introduce a novel capability for LLMs, the \fullmethodname ability, which allows LLMs to recognize errors in their outputs in real-time and correct them simultaneously, thereby generating a more reliable answer. We propose a method for constructing training data for this capability based on wrong step sampling and build the \datasetname based on MetaMathQA. We applied SFT on both the generalist and math-specialized LLMs using the proposed \datasetname, achieving consistent and stable improvements on multiple mathematical benchmarks. Our work demonstrates that LLMs can possess a \fullmethodname ability. There are still areas for improvement in our work, such as generating diverse steps when sampling error steps, examining the quality of reflection and improvement, etc. In the future, we will continue on this work to increase the accuracy during the self-correction process and extend this method to more reasoning tasks, truly generalizing this ability and building a better LLM reasoner.

\section{Acknowledgments}
This work is supported by National Natural Science Foundation of China (No. 62436007), the Key Research and Development Program of Zhejiang Province, China (No. 2023C01152) and ZJU Kunpeng\&Ascend Center of Excellence.

\bibliography{aaai25}

\clearpage

\appendix

\onecolumn
\section{Appendix A. Prompts}
\begin{figure*}[h!]
\centering
\includegraphics[width=0.92\textwidth]{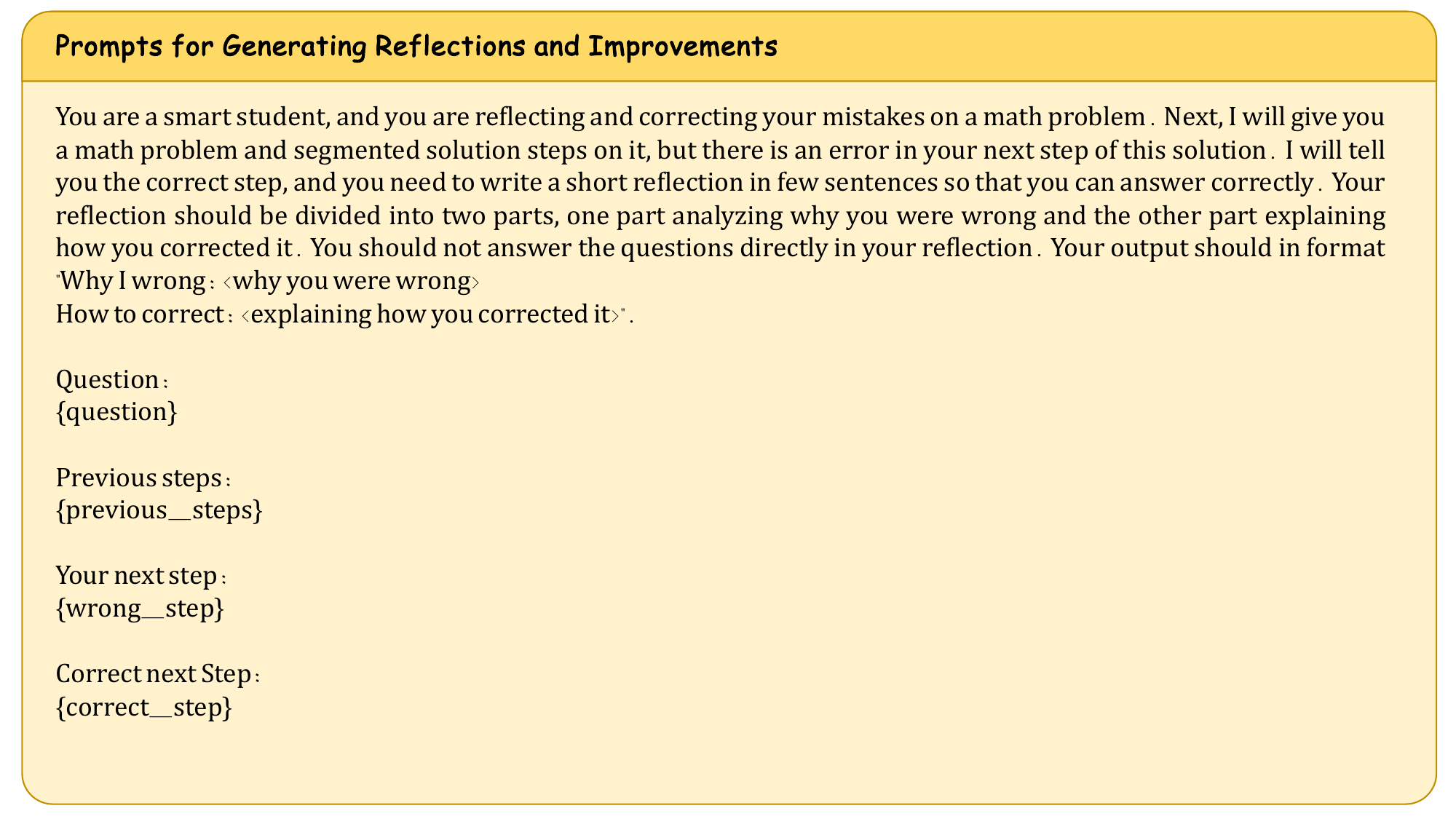} %
\caption{The prompts we used to generate reflection and improvements for \datasetname.}
\label{fig4}
\end{figure*}

In our paper, we employ Meta-Llama-3-70B-Instruct~\citep{metaai2024introducing} for generating reflection and improvement contents based on self-correction. Figure~\ref{fig4} showcases the prompts we utilized.

\end{document}